\documentclass[lettersize,journal]{IEEEtran}
\usepackage{amsmath,amsfonts}
\usepackage{algorithmic}
\usepackage{algorithm}
\usepackage{array}
\usepackage[caption=false,font=normalsize,labelfont=sf,textfont=sf]{subfig}
\usepackage{textcomp}
\usepackage{stfloats}
\usepackage{multirow}
\usepackage{url}
\usepackage{verbatim}
\usepackage{graphicx}
\usepackage{cite}
\usepackage{xcolor}
\usepackage{bm}
\usepackage{amstext}
\usepackage{times}
\usepackage{epsfig}
\usepackage{amsfonts}
\usepackage{amssymb}
\usepackage{color}
\usepackage{booktabs}
\usepackage{bbding}
\usepackage{pifont}
\begin{document}

\title{Fusion-S2iGan: An Efficient and Effective Single-Stage Framework for Speech-to-Image Generation}

\author{Zhenxing Zhang and Lambert Schomaker
\thanks{Zhenxing Zhang and Lambert Schomaker are with the Bernoulli Institute, University of Groningen, Groningen 9747, The Netherlands (e-mail: z.zhang@rug.nl; l.r.b.schomaker@rug.nl.)}
}

\markboth{Journal of \LaTeX\ Class Files,~Vol.~14, No.~8, August~2021}%
{Shell \MakeLowercase{\textit{et al.}}: A Sample Article Using IEEEtran.cls for IEEE Journals}

\IEEEpubid{0000--0000/00\$00.00~\copyright~2021 IEEE}

\maketitle

\begin{abstract}
The goal of a speech-to-image transform is to produce a photo-realistic picture directly from a speech signal. Recently, various studies have focused on this task and have achieved promising performance. However, current speech-to-image approaches are based on a stacked modular framework that suffers from three vital issues: 1) Training separate networks is time-consuming as well as inefficient and the convergence of the final generative model strongly depends on the previous generators; 2) The quality of precursor images is ignored by this architecture; 3) Multiple discriminator networks are required to be trained. To this end, we propose an efficient and effective single-stage framework called Fusion-S2iGan to yield perceptually plausible and semantically consistent image samples on the basis of given spoken descriptions. Fusion-S2iGan introduces a visual+speech fusion module (VSFM), constructed with a pixel-attention module (PAM), a speech-modulation module (SMM) and  a weighted-fusion module (WFM), to inject the speech embedding from a speech encoder into the generator while improving the quality of synthesized pictures. 
Fusion-S2iGan spreads the bimodal information over all layers of the generator network to reinforce the visual feature maps at various hierarchical levels in the architecture. 
We conduct a series of experiments on four benchmark data sets, i.e., CUB birds, Oxford-102, Flickr8k and Places-subset. The experimental results demonstrate the superiority of the presented Fusion-S2iGan compared to the state-of-the-art models with a multi-stage architecture and a performance level that is close to traditional text-to-image approaches.
\end{abstract}

\section{Introduction}
\IEEEPARstart{T}{he} task of speech-to-image generation aims to automatically yield photo-realistic and semantically consistent photographs directly from given speech signals. In recent years, this topic has drawn rapidly growing interest from multidisciplinary communities. It can be potentially used in a wealth of real-world applications, such as creating novel, visually interesting photos providing ideas for visual artists, photograph editing according to the spoken description, generating new data in machine learning \cite{wang2020data}, when augmentation is needed in training, e.g., classifier, and helping disabled persons produce pictures.

The availability of a speech-to-image transform would allow for an AI system to check the visual implications of a spoken narrative. Consider, for example, the `telephone test', where a human describes a 5-image cartoon or comic strip by speech and the system needs to provide its visual associations to reconstruct the original image input. Incongruencies between original and reconstruction may be a useful source of information for continuous learning, both in terms of visual and textual congruence. As an example, a spoken original cartoon description can be visualized by the speech-to-image GAN, its output can be captioned  \cite{zhong2020comprehensive} and this text can be compared to the recognized spoken description. From the textual discrepancies a loss can be computed that is used for training and improving consistency in a cyclic manner.

\IEEEpubidadjcol
Moreover, researching speech-to-image synthesis may help cognitive-science researchers to investigate how humans gradually understand the world. Several studies \cite{polka1994developmental, werker1984cross, gervain2010speech} reveal that human infants know many aspects of their native language by perceiving and analyzing the features of the speech signal such as consonants and vowels. More significantly, infants are able to understand the meanings of numerous ordinary nouns through daily experience with language before they are capable of reading \cite{bergelson20126}. For example, infants will attend to the banana present in their visual field when hearing `Look, a banana!' from their parents. This illustrates that strong connections between vision and speech signals are being created at this early age and infants acquire the capability of discovering the semantic relationships between the spoken words and the visible objects. 
We think that research on synthetic-image generation directly from speech audio may provide a way of looking into the problems that the human brain is confronted with when forming a visual association on the basis of speech input. Note that in this task, an intermediate textual representation can be avoided (learning to read text happens at a much later stage, in children). In traditional AI, the abstract symbolic nature of text is considered to be an essential property and the pinnacle of intelligent information processing. However, the assumption that decoded speech (`recognized text') is a necessity for image generation has not been proven. From the point of view of deep machine learning, there is no strict necessity to rely on an intermediate symbolic text code. Vectorial representations may also be suitable for abstract representations, as has been shown extensively in recent years~\cite{mikolov2013efficient}.

The task of speech-to image generation has a variety of advantages over text-to-image synthesis \cite{zhang2021dtgan, zhang2021divergan}, whose goal is to produce high-resolution image samples semantically aligning with input natural-language descriptions. 

Firstly, spoken language may play a more advanced role in controlling and directing an image-generation process than the text modality. Compared to encoded text, speech signals contain, in addition to the pertinent phonemes, rich characteristics that can be roughly divided into timbre, prosody, rhythm, intonation, stressed components, etcetera \cite{sridhar2008exploiting, du2021phone}. Prosody allows speakers to express the attitude and emotional states, providing key non-verbal cues that help the listener to understand the spoken description \cite{obin2015symbolic}. For instance, when a person says: `There is a fire' in a neutral or falling intonation, it is probable that the fire is not big. However, if the speaker says the same sentence in a sharp intonation, then it may be more likely that this is a case of a large, risky fire. The example shows that text alone may not be sufficient to convey subtleties of the speaker's intention. It is possible to introduce non-verbal cues, e.g., emotional features, for conditional-image synthesis. Incorporating prosody information into the design of picture-generation systems will also be very beneficial for their applications, where users can produce desired photographs by adjusting intonation. Since current text-to-speech synthesis models with deep-learning techniques allow for prosody modulation \cite{du2021phone}, synthesizing pictures on the basis of emotional features and linguistic information of speech signals should be feasible using such produced training data.   

Secondly, the auditory modality is the most commonly used and natural way for humans to communicate information with each other in daily life, while written-form language is more slow and elaborate. 
For these reasons, compact speech-to-image generation systems may become interesting in comparison to a pipeline of a speech-recognizer followed by a standard text-to-image algorithm. 

Thirdly, there are about 3,500 languages lacking orthography or written form \cite{wang2021generating}, which makes it impossible to train a text-to-image synthesis model for these languages and thus text-to-image systems cannot benefit these populations. Since acquiring speech signals is relatively easy and not limited by languages, an adequate speech-to-image generation architecture can be designed for these populations. This may be interesting in language research in the humanities. The presence of such tools would allow for an immediate inspection of the correspondence between spoken description and the generated image, by subjects and scholars.

In summary, spoken descriptions are suitable to explicitly and accurately describe an image using linguistic and emotional information, and researching speech-to-image synthesis has practical and scientific implications.  Nevertheless, speech-to-image generation remains a considerably challenging cross-modal task. Different from textual descriptions, a speech signal is not discrete but continuous, lacking regular breaks between words, i.e., the spaces in the written-form text. This characteristic will make it difficult for the model to grasp the linguistic information of spoken captions of photographs and pictures while learning the corresponding embeddings. Besides, spoken descriptions contain not only the linguistic content but also the non-verbal cues, e.g., emotion, intent and attitude of a speaker, which are significant for generated samples. These features present another crucial challenge for a speech-to-image generation system: \textit{How to effectively inject these features into a neural network which has a large 2D color image as its output?} Note that the common speech-text recognition step is not necessary when using speech-audio descriptions to image generation and emotional features in speech signals would additionally be lost.
Unfortunately, there are as yet no suitable data sets with emotional speech in different degrees, describing a corresponding image in each sample. However, as a first step in the direction of exploiting the potential of speech-audio we will focus on non-emotional speech-to-image transforms.

We are not the first to attempt to translate speech signals into high-resolution pictures directly. For example, Li et al. \cite{li2020direct} introduced a multi-stage speech-to-image GAN architecture to produce photo-realistic pictures semantically correlated with the input spoken description. In order to better capture the linguistic information, the researchers adopted a well-trained image encoder as a `teacher' to train the speech encoder from scratch. Wang et al. \cite{wang2021generating} designed a new speech-embedding network (SEN) to obtain the speech vector. Furthermore, a relation-supervised densely-stacked generative model is developed to yield high-quality photographs. 
\begin{figure}[t]
  \begin{minipage}[b]{1.0\linewidth}
  \centerline{\includegraphics[width=85mm]{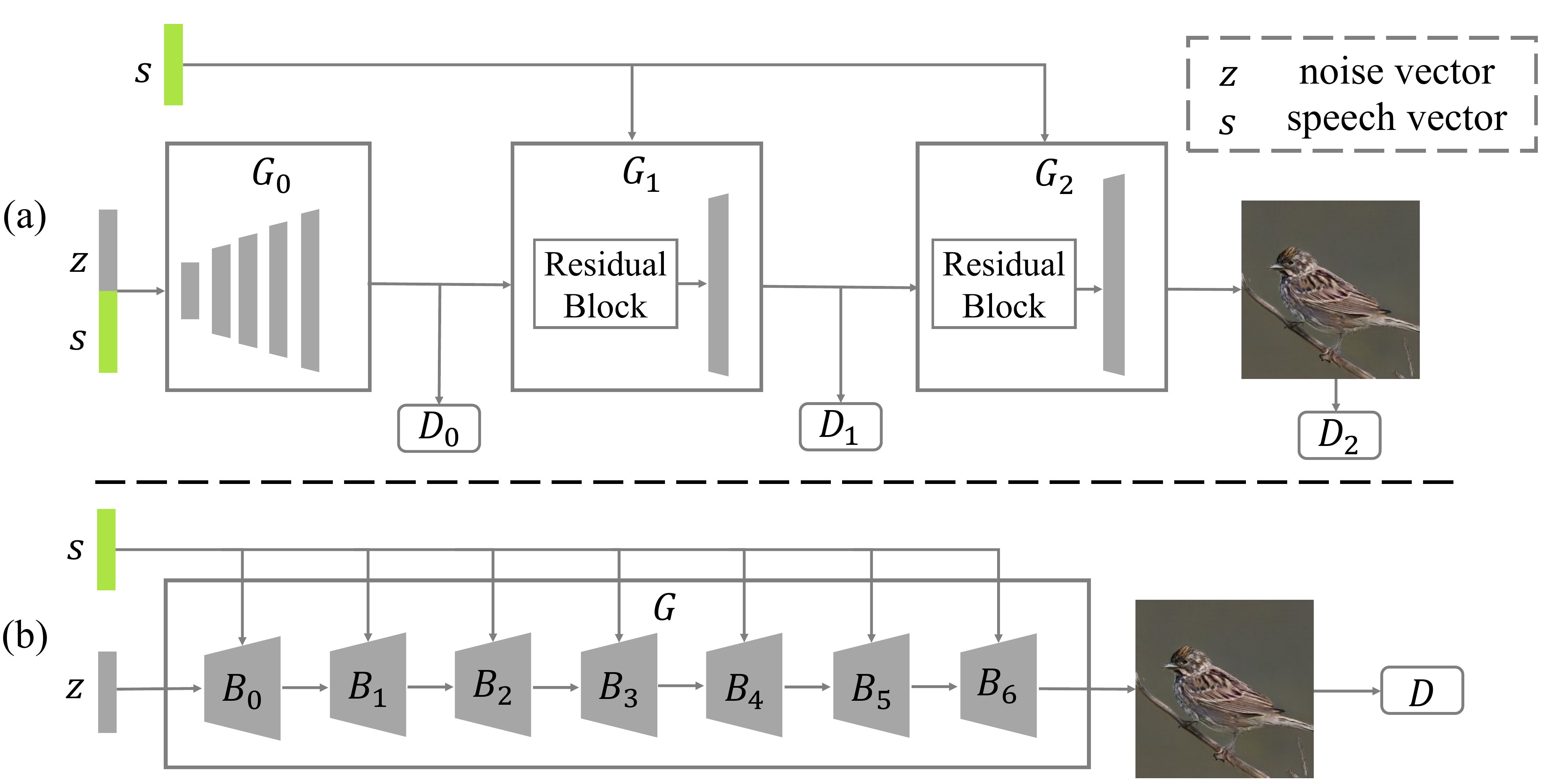}}
  \end{minipage}
  \caption{The comparison between the stacked framework and the proposed architecture. The multi-stage pipeline (a) entails training separate generators to obtain high-quality samples. The presented Fusion-S2iGan (b) is capable of producing visually plausible pictures only employing a single generator/discriminator pair. In (a), $G_{0}$-$G_{2}$ are generators and $D_{0}$-$D_{2}$ are discriminators. In (b), $B_{0}$-$B_{6}$ are the dual-residual speech-visual fusion blocks discussed in Section~\ref{s3}, and $G$ and $D$ are the generator network and discriminator network, respectively.}
  \vspace{-0.1in}
  \label{fig0} 
\end{figure}

These speech-to-image GAN models adopt a multi-stage framework (see Fig. \ref{fig0}(a)), where several generators and discriminators are employed to produce visually plausible samples. Although this architecture has now acquired promising results in the speech-to-image generation task, there still exist three vital problems. First, this framework entails training separate networks and is therefore inefficient as well as time-consuming \cite{zhang2021divergan}. Even worse, it is difficult for the final generator to output perceptually realistic photographs when the earlier generator networks do not converge to a global optimum \cite{zhang2021dtgan}. Second, the quality of the outputs of the previous generator networks \cite{zhu2019dm} is ignored by this architecture. Contextual vectors are not used to enhance and modulate the visual feature maps in the generator for precursor images, which comprises up-sampling operations and convolutional layers \cite{zhang2021dtgan}. Third, several discriminator networks are required to be trained.  

In recent years, we have proposed two novel single-stage text-to-image GAN models, i.e., DTGAN \cite{zhang2021dtgan} and DiverGAN \cite{zhang2021divergan}, to address the above-mentioned issues of a multi-stage architecture. Both DTGAN and DiverGAN are capable of adopting a single generator/discriminator pair to produce photo-realistic and semantically correlated image samples on the basis of given natural-language descriptions. In DTGAN, we presented dual-attention models, conditional adaptive instance-layer normalization and a new type of visual loss. In DiverGAN, we extended the sentence-level attention models introduced in DTGAN to word-level attention modules, in order to better control an image-generation process using word features.  Moreover, we proposed to insert a dense layer into the pipeline to address the lack-of-diversity problem present in current single-stage text-to-image GAN models. Inspired by these previous works and in order to overcome the problems of a stacked framework, we introduce a visual+speech fusion module as well as several effective loss functions, which contribute to a new single-stage speech-to-image architecture. 

The contributions of this paper can be summarized as follows: 

$\bullet$ We present a novel effective and efficient single-stage architecture called Fusion-S2iGan (see Fig. \ref{fig0}(b)) for speech-to-image transforms, which is capable of producing high-quality and semantically consistent pictures only using a generator/discriminator pair. 

$\bullet$ We design a visual+speech fusion module (VSFM) to effectively feed the speech information from a speech encoder to the neural network while improving the quality of generated photographs. More importantly, we spread the bimodal information over almost all layers of the generator. This allows for an influence of the speech over features at various hierarchical levels in  the architecture, from crude early features to abstract late features.

$\bullet$ To the best of our knowledge, we are the first to apply (1) the hinge loss, (2) deep attentional multimodal similarity model (DAMSM) loss and (3) matching-aware zero-centered gradient penalty (MA-GP) loss in speech-to-image generation, which are beneficial for the convergence and stability of the generative model. 

$\bullet$ We carry out extensive experiments on four benchmark data sets, i.e., CUB bird \cite{wah2011caltech}, Oxford-102 \cite{nilsback2008automated}, Flickr8k \cite{hodosh2013framing} and Places-subset \cite{zhou2014learning}. The experimental results suggest that the proposed Fusion-S2iGan has the capacity to yield better pictures than current multi-stage speech-to-image GAN models such as StackGAN++ \cite{zhang2018stackgan++}, Li \textit{et al.} \cite{li2020direct} and S2IGAN \cite{wang2021generating}.

$\bullet$ We explore how far can current single-stage text-to-image methods be used for the speech-to-image transform task, which is depicted in Section \ref{e_vsfm}.

The remainder of this paper is organized as follows. Section~\ref{s2} reviews related works. In Section~\ref{s3}, we introduce the architecture of the proposed Fusion-S2iGan in detail. In Section~\ref{s4}, we elaborate on the presented visual+speech fusion module. Experimental settings are discussed in Section~\ref{s5} and results are reported in Section~\ref{s6}, \ref{s7} and \ref{s8}. Section~\ref{s9} draws the conclusions. 

\section{Related Work}
\label{s2}
In this section, research fields related to our work are described, including text-to-image generation and speech-audio-to-image synthesis. 
\subsection{Text-to-Image Generation}
Many text-to-image synthesis approaches are built upon the original conditional generative adversarial network (cGAN) \cite{mirza2014conditional} due to its appealing performance. 
We roughly group them into two categories in terms of the number of the generators and the discriminators they use. 

\subsubsection{Multi-stage models}
Zhang et al. \cite{zhang2018stackgan++} suggested employing several generators and discriminators to boost image quality and semantic relevance while presenting the first multi-stage text-to-image generation framework named StackGAN. StackGAN serves as a strong basis for the future research. Xu et al. \cite{xu2018attngan} proposed to incorporate a spatial-attention module into the design of a stacked architecture, in order to better bridge the semantic gap between vision and language. Qiao et al. \cite{qiao2019mirrorgan} developed MirrorGAN that introduced an image-to-text model to ensure that synthesized pictures are semantically related to given textual descriptions. Zhu et al. \cite{zhu2019dm} built DMGAN where a dynamic-memory module is applied to enhance the image quality in the initial stage. CPGAN \cite{liang2020cpgan} designed a memory structure to parse the produced image in an object-wise manner and introduced a conditional discriminator to promote the semantic alignment of text-image pairs.

\subsubsection{Single-stage methods}
Reed et al. \cite{radford2015unsupervised} were the first to use a single generator/discriminator pair to yield samples on the basis of natural-language descriptions. However, the resolution of generated pictures is limited owing to the unstable training process as well as the lack of an effective structure. Tao et al. \cite{tao2020df} developed a matching-aware zero-centered gradient penalty loss to help stabilize the training and improve the image quality of a single-stage text-to-image GAN model. Zhang et al. \cite{zhang2021dtgan} presented DTGAN, in which dual-attention models, conditional adaptive instance-layer normalization and a new type of visual loss are designed to generate perceptually realistic images only using a single generator/discriminator pair. Zhang et al. \cite{zhang2021divergan} proposed DiverGAN inserting a dense layer into the pipeline to address the lack-of-diversity problem present in current single-stage text-to-image GAN models. Zhang et al. \cite{zhang2022optimized} introduced linear-interpolation and triangular-interpolation techniques to explain the single-stage text-to-image GAN model. Moreover, a ${Good}$/${Bad}$ data set was created to select successfully generated images and corresponding good latent codes. 
\subsection{Speech-audio-to-Image Synthesis}
With the recent rapid advances in generative-adversarial networks (GANs)~\cite{goodfellow2014generative} and conditional GANs (cGANs) \cite{mirza2014conditional}, speech-to-image generation has made promising advances in image quality and semantic consistency when given speech-audio signals as inputs. 
Various studies focused on synthesizing images conditioned on the sound of music. Chen et al. \cite{chen2017deep} made the first attempt to use the cGAN to produce samples on the basis of music audio. 
Hao et al. \cite{hao2018cmcgan} presented a unified architecture (CMCGAN) for audio-visual mutual synthesis. Specifically, CMCGAN incorporated audio-to-visual, audio-to-audio, visual-to-audio and visual-to-visual networks into the pipeline for cyclic consistency and better convenience.

Some publications tried to reconstruct a facial photograph from a short audio segment of speech. Duarte et al. \cite{duarte2019wav2pix} proposed an end-to-end speech-to-face GAN model called Wav2Pix, which has the ability to synthesize diverse and promising face pictures according to a raw speech signal. Oh et al. \cite{oh2019speech2face} developed a reconstructive speech-to-face architecture named Speech2Face that contains a voice encoder and a pre-trained face decoder network. The encoder is used to extract face information from the given speech and the decoder aims to reconstruct a realistic face sample. 

Different from the above approaches, several papers aimed at translating a spoken description of an image into a high-quality picture directly. Li et al. \cite{li2020direct} attempted to apply a multi-stage speech-to-image GAN model to yield perceptually plausible pictures semantically correlated with input speech-audio descriptions. To better acquire the speech embedding, the researchers used a well-trained image encoder as a `teacher' to train the speech encoder from scratch. Wang et al. \cite{wang2021generating} proposed S2IGAN where a speech-embedding network (SEN) was designed to obtain the spoken vector and a matching loss and a distinctive loss were presented to train SEN. In addition, a relation-supervised densely-stacked generative model is introduced to produce high-resolution images.  

This paper focuses on solving the task of speech-to-image synthesis only using a single generator/discriminator pair.

\section{Fusion-S2iGan for speech-to-image generation}
\label{s3}
\begin{figure*}[t]
   \begin{minipage}[b]{1.0\linewidth}
   \centerline{\includegraphics[width=180mm]{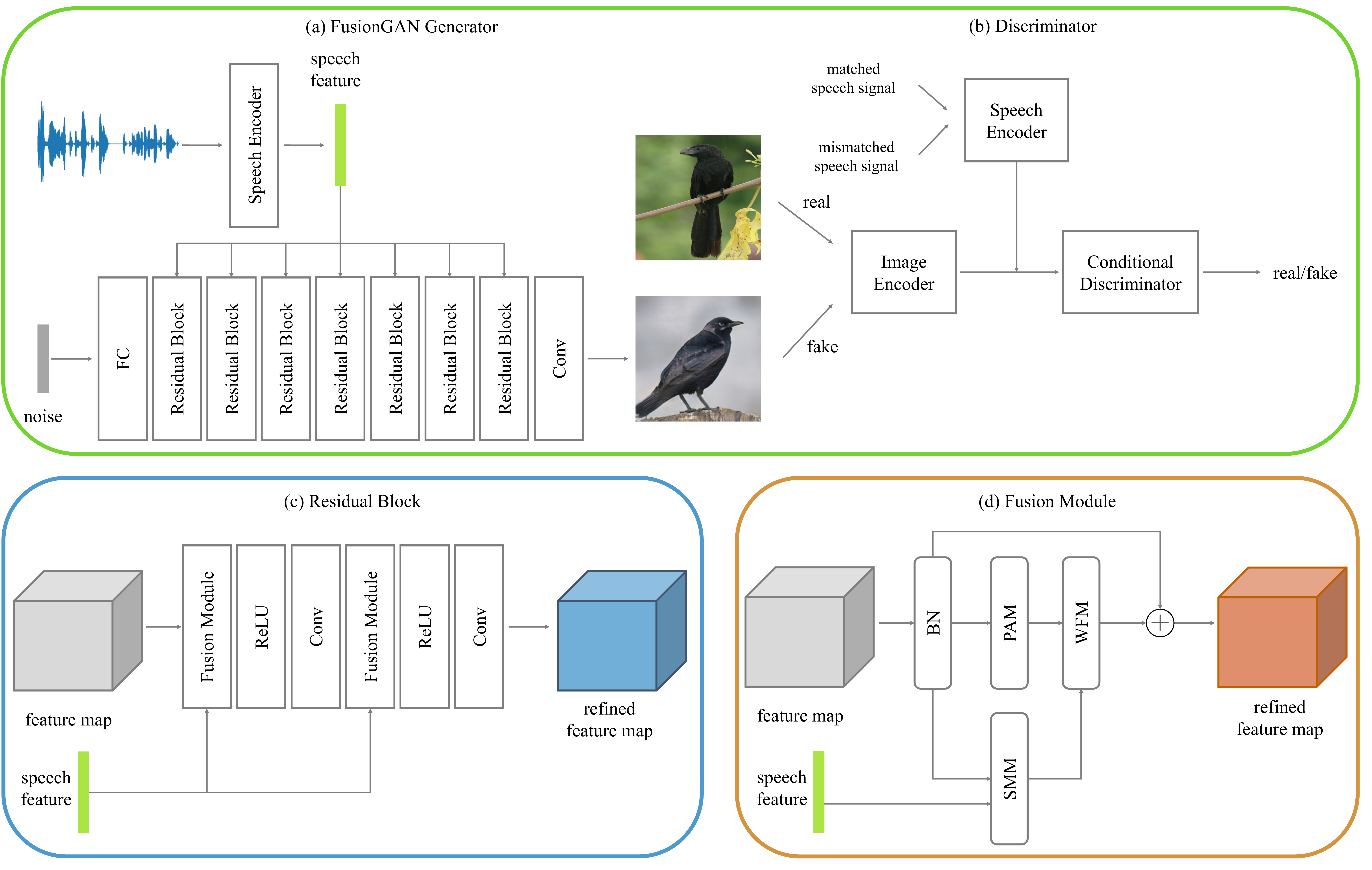}}
   \end{minipage}
   \caption{The overall framework of the proposed Fusion-S2iGan. \textit{FC} is a dense layer, \textit{Conv} is a convolutional layer, \textit{ReLU} is a ReLU activation function and \textit{BN} is a Batch-Normalization operation. Additionally, \textit{Residual Block} and \textit{Fusion Module} refer to the presented dual-residual speech-visual fusion block (see (c)) discussed in Section~\ref{generator} and visual+speech fusion module (see (d)) discussed in Section~\ref{s4}, respectively. Furthermore, \textit{PAM}, \textit{SMM} and \textit{WFM} in (d) represent the pixel-attention module, speech-modulation module and weighted-fusion module, respectively, discussed in Section~\ref{s4}. Note that we do not plot the up-sample layers between \textit{Residual Blocks} in (a) due to the limited space.}
   \vspace{-0.1in}
  \label{architecture} 
\end{figure*}

The overall framework of Fusion-S2iGan for a speech-to-image transform is presented in Fig. \ref{architecture}. The architecture only consists of a generator and a discriminator. Next, we introduce these two components one by one. 
\subsection{Preliminary}
The goal of speech-to-image generation is to yield perceptually plausible samples that are semantically correlated with the linguistic content of input spoken descriptions. Mathematically, let $\{ (I_{i}, S_{i})\}_{i=1}^{n}$ represent a suite of $n$ image-spoken caption pairs for training, where $I_{i}$ indicates an image and $S_{i}=(s_{i}^{1}, s_{i}^{2}, ..., s_{i}^{k})$ refers to a set of $k$ speech-audio descriptions of $I_{i}$. The generator of a speech-to-image GAN model aims to synthesize a high-resolution and semantically related picture $\hat{I}_{i}$ on the basis of a speech signal $s_{i}$ randomly picked from $S_{i}$. In the meantime, the discriminator is trained to separate the real image-speech pair $(I_{i}, s_{i})$ from the fake image-speech pair $(\hat{I}_{i}, s_{i})$.
\subsection{Generator}
\label{generator}
The generator is able to project a speech signal into a photo-realistic and semantically consistent picture, shown in Fig. \ref{architecture}(a). More specifically, the generator network is composed of a dense layer transforming a latent code to the initial feature map and seven dual-residual speech-visual fusion blocks modulating the visual feature map with the spoken vector derived from a speech encoder. The speech encoder is employed to learn the semantic representation and conceptual meaning of a given spoken description that capture the discriminative visual details \cite{haque2021guided}.  

The designed dual-residual speech-visual fusion block (see Fig. \ref{architecture}(c)) contains two effective and efficient visual+speech fusion modules (see Fig. \ref{architecture}(d)) as well as a suite of ReLU activation functions and convolutional layers. Each visual+speech fusion module comprises Batch Normalization (BN) \cite{santurkar2018does}, a pixel-attention module (PAM), a speech-modulation module (SMM) and a weighted-fusion module (WFM). The dual-residual speech-visual fusion block allows us to easily enhance model capacity by effectively increasing the number of layers, while also stabilizing the training process by maintaining more original features than cascade structures. Benefiting from such dual-residual speech-visual fusion blocks, the generator network has the ability to yield high-quality pictures. The process of synthesizing photographs on the basis of a speech signal is formulated as follows:
\begin{align}
&h_{0}=F_{0}(z) \\
&h_{1}=F_{1}^{Dual}(h_{0},s) \\
&h_{i}=F_{i}^{Dual}(h_{i-1}\uparrow,s)  \quad for \quad i=2,3,...,7 \\
&o=G_{c}(h_{7}) 
\end{align}
where $z$ is a latent vector randomly sampled from the normal distribution, $F_{0}$ is a dense layer, $s$ is the speech embedding from a speech encoder, $F_{i}^{Dual}$ is the presented dual-residual speech-visual fusion block and $G_{c}$ is the last convolutional layer. The details of the proposed visual+speech fusion module will be discussed in Section~\ref{s4}.

\subsection{Discriminator}
The architecture of the discriminator network comprises an image encoder, a pre-trained speech encoder and a conditional discriminator network, depicted in Fig. \ref{architecture}(b). To be specific, the image encoder is constructed with one convolutional layer with strides 1 kernel size 3 padding 1 followed by six adaptively successive residual blocks. Each block consists of two convolutional layers, where the first layer with strides 2 kernel size 4 padding 1 is used to reduce the dimension to half of the input feature map and the second one with strides 1 kernel size 3 padding 1 aims to further distill image features. After each convolutional layer, Leaky-ReLU activation \cite{maas2013rectifier} with a slope of 0.2 is utilized to help the training. The number of filters for residual blocks are 64 128 256 512 1024 1024, respectively. In order to stabilize the learning, we incorporate a residual connection into each block. In the process of training, a picture is fed into the image encoder to extract the image features (1024, 4, 4) which are combined with the spoken vector from the speech encoder as the joint embeddings. After that, the joint features are passed through the conditional discriminator to gain the final conditional score, which is utilized to determine whether the input image-spoken caption pair is real or fake. 
\subsection{Objective Function}
An adversarial loss is employed to match generated samples to input speech signals. Inspired by \cite{zhang2021divergan}, we utilize the hinge objective \cite{lim2017geometric} for stable training instead of the vanilla GAN objective. The adversarial loss for the discriminator is defined as follows:
\begin{equation}
\begin{split}
\mathcal{L}_{\text{adv}}^{D}=&\mathbb{E}_{x\sim p_{\text{data}}}\left [\text{max}(0,1-D(x,s)) \right ]\\
&+\frac{1}{2}\mathbb{E}_{x\sim p_{G}}\left [\text{max}(0,1+D(\hat{x},s)) \right ]\\
&+\frac{1}{2}\mathbb{E}_{x\sim p_{\text{data}}}\left [\text{max}(0,1+D(x,\hat{s})) \right ] 
\end{split}
\end{equation}
where $s$ is a given spoken caption, $\hat{s}$ is a mismatched speech-audio description, $x$ is the real image from the distribution $p_{data}$ and $\hat{x}$ is the synthesized sample from the distribution $p_{G}$, 

To enhance the image quality and semantic consistency of produced pictures, we adopt the matching-aware zero-centered gradient penalty (MA-GP) loss \cite{tao2020df} for the discriminator, which applies gradient penalty to real images and input spoken descriptions. The MA-GP Loss is formulated as follows:
\begin{equation}
\mathcal{L}_{\text{M}}=\mathbb{E}_{x\sim p_{\text{data}}}\left [(\left \| \nabla_{x}D(x,s) \right \|_{2}+\left \| \nabla_{s}D(x,s) \right \|_{2})^{p}\right ]
\end{equation}

For the generator, we apply an adversarial loss and a deep attentional multimodal similarity model (DAMSM) loss \cite{xu2018attngan} to train the network. 
\section{Visual+Speech Fusion Module} 
\label{s4}
It is widely known that the semantic relationships between the visual content of photographs and images and the corresponding conditional contexts, e.g., class labels and textual descriptions, play a significant role in an image-generation process. However, this correlation will be more complicated for speech-to-image synthesis, since the speech signal is long and continuous, lacking word boundaries, i.e., the spaces in natural-language descriptions. It is thus very difficult to model the affinities between the linguistic content of spoken descriptions and the visible objects. Moreover, word-level modulation modules fail to fuse spoke information and visual feature maps due to the continuity of speech signals. In this case, a crucial problem is most clearly present for researchers: \textit{how to effectively inject the information from a spoken probe into a neural network which has a large 2D color image as its output?} 

The input speech signal needs to be converted to the global speech vector similar to the sentence embedding \cite{zhang2021dtgan} and then can be made to modulate feature maps in the network. In this section, we develop an effective and efficient visual+speech fusion module (VSFM) to facilitate the visual feature maps with the global speech embedding and yield high-resolution pictures.
\subsection{Overall Architecture}
The framework of the proposed visual+speech fusion module is shown in Fig. \ref{architecture}(d). Given an intermediate feature map $F\in \mathbb{R}^{C\times H\times W}$ and the speech vector $s \in \mathbb{R}^{D}$ from a speech encoder as inputs, the VSFM first applies Batch Normalization (BN) \cite{santurkar2018does} on $F$, acquiring a new feature map $F{}'\in \mathbb{R}^{C\times H\times W}$. This may help stabilize the learning of the conditional generative adversarial network (cGAN) and accelerate the training process. After that, the VSFM refines $F{}'\in \mathbb{R}^{C\times H\times W}$ using the pixel-attention module (PAM) $M_{p}$ and speech-modulation module (SMM) $M_{s}$, respectively. Afterwards, the VSFM effectively fuses their outputs $M_{p}(F{}')\in \mathbb{R}^{C\times H\times W}$ and $M_{s}(F{}', s)\in \mathbb{R}^{C\times H\times W}$ using the weighted-fusion module (WFM) $M_{f}$. Meanwhile, a residual connection is employed to get the final enhanced result $F{}'' \in \mathbb{R}^{C\times H\times W}$. Note that we spread the bimodal information of the VSFM over all layers of the generator network. This allows for an influence of the speech over features at various hierarchical levels in the architecture, from crude early features to abstract late features. The overall process of the VSFM can be formulated as follows:
\begin{align}
&F{}'=BN(F) \\
&F{}''=F{}'+M_{f}(M_{p}(F{}'), M_{s}(F{}', s))
\end{align}
where $BN$ indicates a Batch-Normalization operation. The details of PAM, SMM and WFM will be described in the following subsections.
\subsection{Pixel-Attention Module (PAM)} 
Pictures and photographs consist of visual pixels which are considerably significant for image quality. Learning the long-range contextual dependency of each position of a picture is essential for producing perceptually plausible samples. However, convolutional operations can only grasp local relationships between spatial contexts and thus fail to `see' the entire image field. Hence, the pixel-attention module (PAM) is introduced to effectively model the spatial affinities between visual pixels and enable crucial and informative positions to receive more attention from the generator.  
\begin{figure*}[t]
   \begin{minipage}[b]{1.0\linewidth}
   \centerline{\includegraphics[width=180mm]{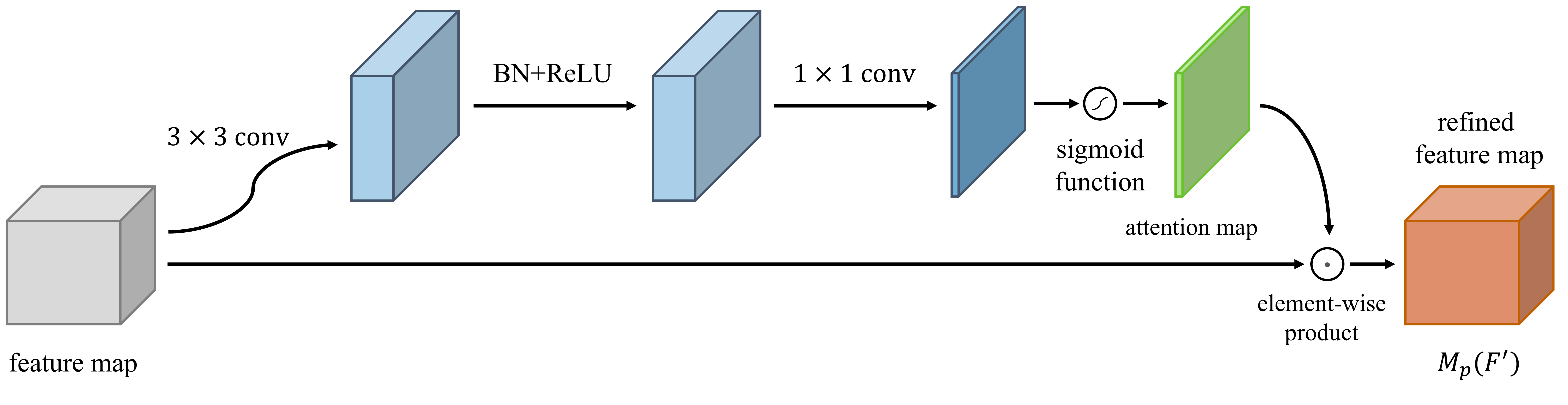}}
   \end{minipage}
   \caption{Overview of the introduced pixel-attention module, which aims to assign larger weights to discriminative and informative pixels. \textit{BN} and \textit{ReLU} refer to Batch Normalization and a ReLU activation function, respectively. \textit{$1\times1$ conv} and \textit{$3\times3$ conv} indicate the $1\times1$ and $3\times3$ convolutional operation, respectively.}
   \vspace{-0.1in}
  \label{PAM} 
\end{figure*}

Fig. \ref{PAM} illustrates the process of the pixel-attention module (PAM). 
For a feature map $F{}'\in \mathbb{R}^{C\times H\times W}$, we first feed it into a $3\times 3$ convolutional layer followed with BN and a ReLU function to reduce the channel dimension to $\mathbb{R}^{C/r\times H\times W}$. This may integrate and strength the visual feature map across the channel and spatial directions. Subsequently, we use a $1\times 1$ convolutional layer followed with the sigmoid function to process the features to obtain the pixel-attention map $PA \in \mathbb{R}^{1\times H\times W}$. Mathematically,
\begin{equation}
PA=\sigma(f_{1}^{1\times 1}(ReLU(BN(f_{0}^{3\times 3}(F{}')))))
\end{equation}
where $f$ is a convolutional layer, $ReLU$ is the ReLU function and $\sigma$ is the sigmoid function. We conduct a matrix multiplication between the original feature map $F{}'\in \mathbb{R}^{C\times H\times W}$ and the spatial-attention map $PA \in \mathbb{R}^{1\times H\times W}$ to acquire the refined result. Specifically,  
\begin{equation}
M_{p}(F{}')=F{}'\odot PA
\end{equation}
where $\odot$ is the element-wise multiplication. 
\subsection{Speech-Modulation Module (SMM)}
An adequate modulation method needs to be developed to ensure the semantic consistency and quality of synthesized photographs \cite{zhang2021divergan}. To this end, the speech-modulation module (SMM) is introduced to inject the features derived from the speech signal into the network at the proper points in the network architecture. 

Inspired by \cite{tao2020df}, SMM facilitates the visual feature map using detailed linguistic cues captured from the speech embedding $s$. To be specific, we adopt two dense layers to project $s$ into the linguistic cues $WA \in \mathbb{R}^{C\times 1\times 1}$ and $BA  \in \mathbb{R}^{C\times 1\times 1}$. After that, $WA$ and $BA$ are employed to scale and shift $F{}'$. The process of SMM can be defined as follows:
\begin{align}
&WA=MLP(ReLU(MLP(s))) \\
&BA=MLP(ReLU(MLP(s))) \\
&M_{s}(F{}')=F{}'\odot WA+ BA
\end{align}
where $MLP$ is a fully-connected perceptron layer and $M_{s}(F{}')$ is the output from SMM. 
\subsection{Weighted-Fusion Module (WFM)}
We do not simply fuse the attended and enhanced visual feature maps from PAM and SMM using an addition or element-wise multiplication, since pixels and speech signals refer to two very different modalities and need to be combined in a more advanced manner. Here, we propose an efficient weighted-fusion module (WFM) to highlight the discriminative and significant regions in an adaptive manner and produce high-quality image samples.
\begin{figure*}[t]
   \begin{minipage}[b]{1.0\linewidth}
   \centerline{\includegraphics[width=180mm]{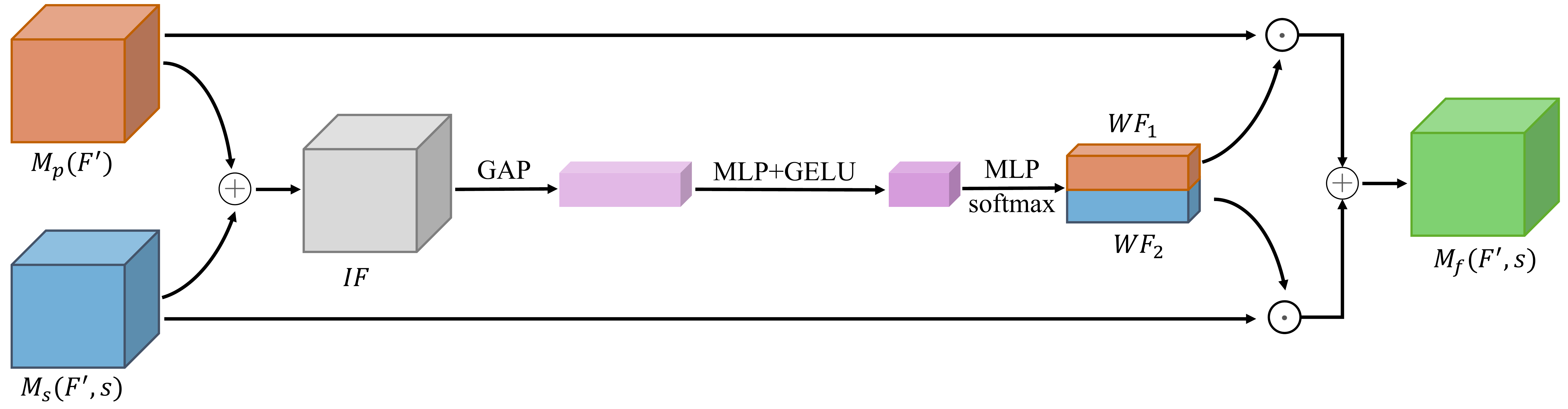}}
   \end{minipage}
   \caption{Overview of the proposed weighted-fusion module, which effectively combines the outputs from the pixel-attention module (PAM) and speech-modulation module (SMM) in an adaptive manner. $M_{p}(F{}')$ and $M_{s}(F{}', s)$ denote the result of PAM and SMM, respectively. \textit{GAP}, \textit{MLP} and \textit{GELU} represent the global-average pooling, fully-connected perceptron layer and Gaussian Error Linear Unit (GELU) \cite{hendrycks2016gaussian}, respectively. $WF_{1}$ and $WF_{2}$ are the channel-aware weight matrix. $M_{f}(F{}', s)$ is the final refined result.}
   \vspace{-0.1in}
  \label{WFM} 
\end{figure*}

The detailed structure of the weighted-fusion module (WFM) is depicted in Fig. \ref{WFM}. For the outputs $M_{p}(F{}')$ and $M_{s}(F{}', s)$ from PAM and SMM, respectively, we perform an element-wise addition between them as a first step, acquiring an intermediate feature map $IF\in \mathbb{R}^{C\times H\times W}$. In the next step, $IF$ is used to compute the final weights for both $M_{p}(F{}')$ and $M_{s}(F{}', s)$. More specifically, we employ global-average pooling (GAP) to process $IF$ to aggregate holistic and discriminative information, thereby obtaining a channel-feature vector $IF{}'\in \mathbb{R}^{C\times 1\times 1}$. Subsequently, we feed $IF{}'$ into two fully-connected perceptron layers, in which the first one is to compress and integrate the channel features and the other aims to recover the channel dimension and capture the semantic importance of the image-attention mask and the speech-modulation module at the level of the channels. After that, we apply a softmax function across the channel dimension to get the contextually channel-aware weight matrix $WF \in \mathbb{R}^{C\times 2}$. Afterwards, $WF$ is split into two separate channel-wise weight matrices $WF_{1} \in \mathbb{R}^{C\times 1}$ and $WF_{2} \in \mathbb{R}^{C\times 1}$. Mathematically,   
\begin{align}
&IF=M_{p}(F{}')+M_{s}(F{}', s) \\
&IF{}'=GAP(IF)\\
&WF=softmax(MLP(GELU(MLP(IF{}')))) \\
&WF=[WF_{1}; WF_{2}]
\end{align}
where $GAP$ is the global-average pooling and $GELU$ is the Gaussian Error Linear Unit (GELU) \cite{hendrycks2016gaussian}. $WF_{1}$ and $WF_{2}$ are resize into  $\mathbb{R}^{C\times 1\times 1}$ and applied to $M_{p}(F{}')$ and $M_{s}(F{}', s)$ to get the final fusion result. It is denoted as follows:
\begin{equation}
M_{f}(F{}', s)=WF_{1}\odot M_{p}(F{}') + WF_{2}\odot M_{s}(F{}', s)
\end{equation}
\section{Experimental Settings}
\label{s5}
\subsection{Data Sets}
To evaluate the proposed Fusion-S2iGan, we perform extensive experiments on two synthesized spoken caption-image data sets and two real spoken caption-image data set, which are employed by Li \textit{et al.} \cite{li2020direct} and S2IGAN \cite{wang2021generating}.

$\bullet$ \textbf{CUB bird \cite{wah2011caltech}.} The CUB data set includes a total of 11,788 images, which is divided into 8,855 training pictures and 2,933 testing pictures. Each picture is accompanied by 10 natural-language descriptions. To evaluate the task of speech-to-image generation, Li \textit{et al.} \cite{li2020direct} and S2IGAN \cite{wang2021generating} transform textual descriptions to spoken captions utilizing a text-to-speech method such as Tacotron 2 \cite{shen2018natural}.   

$\bullet$ \textbf{Oxford-102 \cite{nilsback2008automated}.} The Oxford-102 data set is composed of 8,189 pictures, in which 5,878 pictures belong to the training set and the other 2,311 pictures are used for testing. Each picture contains 10 textual descriptions, which are transformed to speech signals in the same way as the CUB data set. 

$\bullet$ \textbf{Flickr8k \cite{hodosh2013framing}.} The Flickr8k data set is a more challenging data set comprising 8,000 scene pictures and each image is paired with 5 real spoken descriptions collected by \cite{harwath2015deep}. We split the Flickr8k data set according to S2IGAN \cite{wang2021generating}. 

$\bullet$ \textbf{Places-subset \cite{zhou2014learning}} The Places-subset data set is a subset of the Places Audio Caption data set \cite{harwath2018jointly, harwath2016unsupervised}, which encompasses real spoken captions of pictures from the Places 205 data set \cite{zhou2014learning}. It contains a total of 13,803 image-spoken caption pairs belonging to 7 categories, which are divided into 10,933 training paired data and 2,870 testing paired data.
\subsection{Implementation Details}
For the speech encoder, following the structure of S2IGAN \cite{wang2021generating}, we use two 1-D convolution blocks, two bidirectional gated recurrent units (GRUs) and a self-attention module to obtain the speech embedding. To better capture the semantic representation of spoken descriptions on the Places-subset data set, we replace the Inception-V3 \cite{szegedy2016rethinking} pre-trained on ImageNet \cite{russakovsky2015imagenet} in S2IGAN with the ResNet \cite{he2016deep} trained on Places 205 \cite{zhou2014learning}. The dimension of the speech vector is set to 1024. In the experiments, we utilize the Adam optimizer \cite{kingma2014adam} with $\beta_{1}=0.0$ and $\beta_{2}=0.9$ to train the networks. Furthermore, we follow the two timescale update rule (TTUR) \cite{heusel2017gans} and set the learning rates for the generator and the discriminator to 0.0001 and 0.0004, respectively. We set the batch size to 32. We implement Fusion-S2iGan adopting PyTorch~\cite{paszke2019pytorch}. All the experiments are conducted on a single NVIDIA Tesla V100
GPU (32 GB memory).
\subsection{Evaluation Metrics}
We verify the effectiveness of Fusion-S2iGan by computing the following four extensively employed evaluation metrics:

$\bullet$ \textbf{Inception score (IS) \cite{salimans2016improved}}. The IS is obtained by computing the KL divergence between the conditional class distribution and the marginal class distribution. The synthesized pictures are divided into multiple groups and the IS is calculated on each group of photographs, then the average and standard deviation of the score are reported. Higher IS demonstrates better quality and diversity among the generated images \cite{zhang2021divergan}.

$\bullet$ \textbf{Fr\'echet inception distance (FID) \cite{szegedy2016rethinking}.} The FID calculates the Fr\'echet distance between the distribution of generated images and the distribution of true data. A lower FID score means that the generated pictures are closer to the corresponding real pictures. 

$\bullet$ \textbf{Mean average precision (mAP) and R(ecall)@50 \cite{wang2021generating}.} The mAP and R(ecall)@50 are speech-image retrieval metrics, introduced by S2IGAN to measure the semantic relevancy for the synthesized and real speech data sets, respectively. Higher mAP or R@50 suggest better semantic consistency. 

It should be noted that standard deviation on performances cannot be given for all metrics reported (FID and mAP). This is due to the fact that we wanted the measurements to be comparable with other studies. However, given the size of the test sets: $N=30k$ for the CUB data set and $N=10k$ for the Oxford-102 data set, we expect the measurements to be fairly reliable. As a comparative illustration: In case of a measured accuracy of 0.80 and a data set of $N=30k$, at $\alpha=0.01$, the confidence band would span 0.794 to 0.806, i.e., a deviation of just 0.75\%.
\section{Results on the synthesized speech datasets}
\label{s6}
\subsection{Quantitative Results}
We compare Fusion-S2iGan with previous single-stage \cite{tao2020df, zhang2020dtgan, zhang2021divergan} and multi-stage \cite{zhang2018stackgan++, xu2018attngan, qiao2019mirrorgan, li2019controllable, yin2019semantics, zhu2019dm, li2020direct, wang2021generating} cGAN-based methods in text-to-image generation and speech-to-image transforms on the CUB and Oxford-102 data sets. The IS, FID and mAP of Fusion-S2iGan and other compared approaches on the CUB and Oxford-102 data sets are shown in Table \ref{t1}. Note that the reported scores in Table \ref{t1} are based on the results presented in these publications. We can see that Fusion-S2iGan achieves the best scores on speech-to-image synthesis, significantly improving the IS from 4.29 to 4.82 on the CUB data set and from 3.69 to 3.81 on the Oxford-102 data set, reducing the FID from 14.50 to 13.74 on the CUB data set and from 48.64 to 40.08 on the Oxford-102 data set and increasing the mAP from 9.04 to 11.49 on the CUB data set and from 13.40 to 17.72 on the Oxford-102 data set. Notably, Fusion-S2iGan performs better in speech-to-image generation than both DM-GAN \cite{zhu2019dm} in text-to-image generation on the CUB data set, as seen from an improvement of the IS from 4.75 to 4.82 and DTGAN \cite{zhang2021dtgan} doing text-to-image synthesis on the Oxford-102 data set, here improving the IS from 3.77 to 3.81. The experimental results suggest that Fusion-S2iGan is capable of yielding perceptually plausible pictures with higher quality and better diversity than state-of-the-art speech-to-image transform models and many text-to-image approaches. 
\begin{table}
\caption{The IS, FID and mAP of previous text-to-image and speech-to-image approaches and Fusion-S2iGan on the CUB and Oxford-102 data sets. The best results of speech-to-image generation are in bold.}
\begin{center}
\scalebox{0.88}{
\begin{tabular}{c c c c c c}
\midrule
Datasets & Methods & Input & IS $\uparrow$ & FID $\downarrow$ & mAP $\uparrow$\\
\midrule
\multirow{16}*{CUB} 
& StackGAN++ \cite{zhang2018stackgan++} & text & 4.04$\pm$0.05 & 26.07 & 7.01\\
& AttnGAN \cite{xu2018attngan} & text & 4.36$\pm$0.03 & 23.98 & $-$\\
& MirrorGAN \cite{qiao2019mirrorgan} & text & 4.56$\pm$0.05 & $-$ & $-$\\
& ControlGAN \cite{li2019controllable} & text & 4.58$\pm$0.09 & $-$ & $-$\\
& SDGAN \cite{yin2019semantics} & text & 4.67$\pm$0.09 & $-$ & $-$\\
& DM-GAN \cite{zhu2019dm} & text & 4.75$\pm$0.07 & 16.09 & $-$\\
& DF-GAN \cite{tao2020df} & text & 4.86$\pm$0.04 & 19.24 & $-$\\
& DTGAN \cite{zhang2020dtgan} & text & 4.88$\pm$0.03 & 16.35 & $-$ \\
& DiverGAN \cite{zhang2021divergan} & text & 4.98$\pm$0.06 & 15.63 & $-$ \\
\cmidrule(r){2-6}
& Classifier-based \cite{wang2021generating} & speech & 3.68$\pm$0.04 & 43.76 & $-$\\
& Li \textit{et al.} \cite{li2020direct} & speech & 4.09$\pm$0.04 & 18.37 & $-$\\
& StackGAN++ \cite{zhang2018stackgan++} & speech & 4.14$\pm$0.04 & 18.94 & 8.09\\
& S2IGAN \cite{wang2021generating} & speech & 4.29$\pm$0.04 & 14.50 & 9.04\\
\cmidrule(r){2-6}
& \textbf{Fusion-S2iGan} & speech & \textbf{5.06$\pm$0.09} & \textbf{13.09} & \textbf{12.12}\\
\midrule
\multirow{11}*{Oxford-102}
& StackGAN++ \cite{zhang2018stackgan++} & text & 3.26$\pm$0.01 & 48.68 & $-$ \\
& DF-GAN \cite{tao2020df} & text & 3.71$\pm$0.06 & $-$ & $-$ \\
& DTGAN \cite{zhang2020dtgan} & text & 3.77$\pm$0.06 & $-$ & $-$ \\
& DiverGAN \cite{zhang2021divergan} & text & 3.99$\pm$0.05 & $-$ & $-$\\
\cmidrule(r){2-6}
& Classifier-based \cite{wang2021generating} & speech & 3.30$\pm$0.06 & 64.75 & $-$ \\
& Li \textit{et al.} \cite{li2020direct} & speech & 3.23$\pm$0.05 & 54.76 & $-$ \\
& StackGAN++ \cite{zhang2018stackgan++} & speech & 3.69$\pm$0.08 & 54.33 & 12.18 \\
& S2IGAN \cite{wang2021generating} & speech & 3.55$\pm$0.04 & 48.64 & 13.40\\
\cmidrule(r){2-6}
& \textbf{Fusion-S2iGan} & speech & \textbf{3.81$\pm$0.08} & \textbf{40.08} & \textbf{17.72}\\
\midrule
\end{tabular}}
\end{center}
\label{t1}
\vspace{-0.1in}
\end{table}
\begin{figure*}[t]
   \begin{minipage}[b]{1.0\linewidth}
   \centerline{\includegraphics[width=180mm]{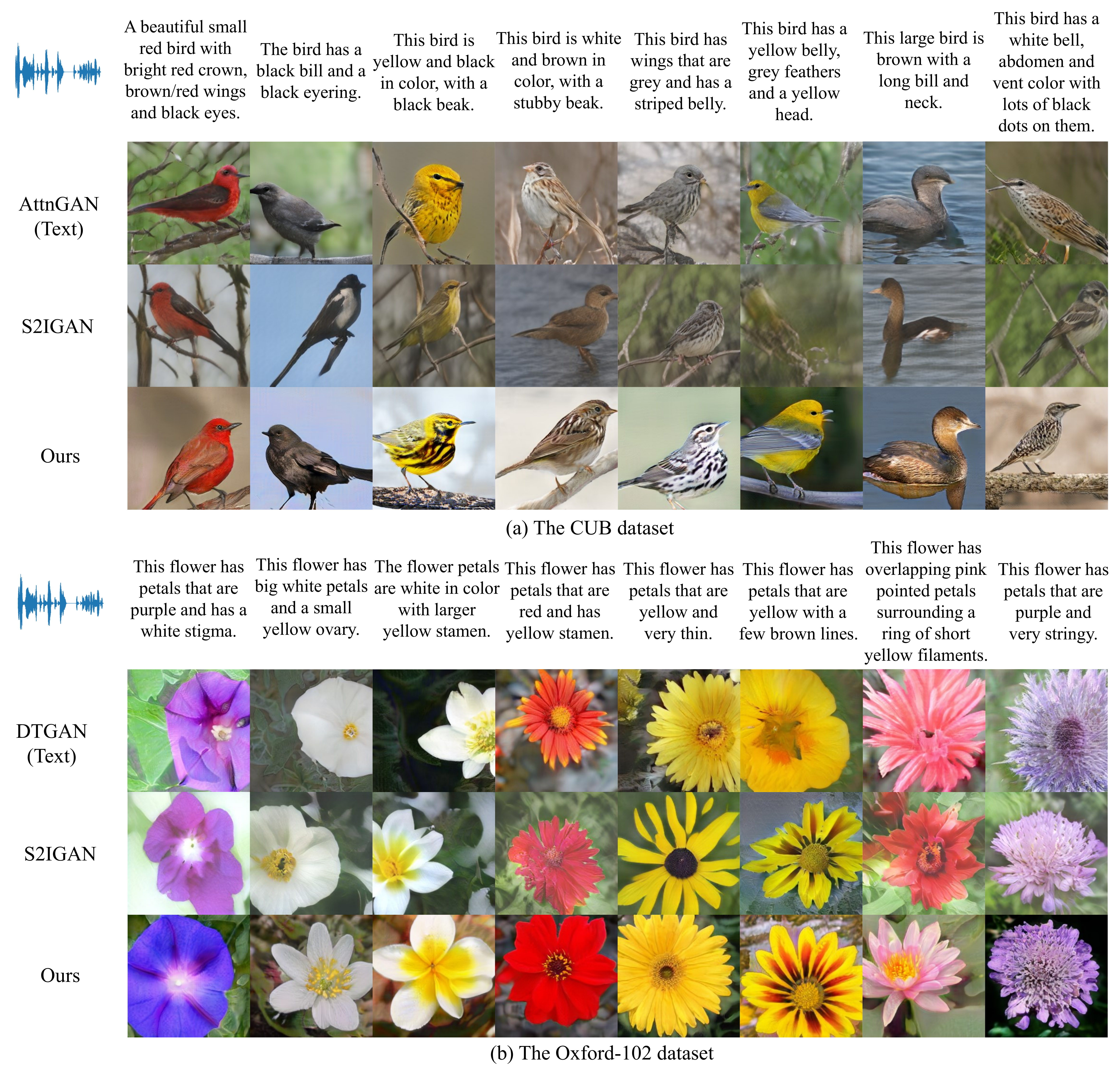}}
   \end{minipage}
   \caption{Qualitative comparison of AttnGAN \cite{xu2018attngan}, DTGAN \cite{zhang2021dtgan} conditioned on the textual descriptions, S2IGAN \cite{wang2021generating} and Fusion-S2iGan on the basis of the speech signals on the CUB and Oxford-102 data sets. The spoken descriptions are shown above the images.}
   \vspace{-0.1in}
  \label{q1} 
\end{figure*}
\subsection{Qualitative Results}
In addition to quantitative experiments, we carry out qualitative comparison on the CUB and Oxford-102 data sets, which is presented in Fig. \ref{q1}. Note that in order to visualize the synthesized images of S2IGAN \cite{wang2021generating}, we utilize the public code \footnote{\url{https://github.com/xinshengwang/S2IGAN}} to train the network from scratch due to the lack of available pre-trained models. It was noted that the contrast of the generated images was dull for the S2IGAN code, as given. By enhancing brightness and contrast, this problem could be solved. However, we found this is only needed for human evaluation: On 20 selected images  \footnote{\url{https://zenodo.org/record/7014899#.YwN-xHZByUk}}, the IS score for S2IGAN was the same for the raw (greyish) output and the enhanced output samples. 
%
%
As can be seen in Fig. \ref{q1}(a), the shape of birds produced by AttnGAN \cite{xu2018attngan} and S2IGAN \cite{wang2021generating} ($2^{nd}$, $3^{rd}$, $6^{th}$ and $8^{th}$ column) is strange, the image details are lost ($2^{nd}$, $3^{rd}$, $5^{th}$, $6^{th}$ and $7^{th}$ column) and the backgrounds are blurry ($1^{st}$, $3^{rd}$, $6^{th}$ and $8^{th}$ column). However, Fusion-S2iGan yields more clear and photo-realistic samples than AttnGAN and S2IGAN, which demonstrates the effectiveness of our architecture. For example, the birds synthesized by Fusion-S2iGan have a more clear shape and richer color distributions compared to AttnGAN and S2IGAN in the $2^{nd}$, $3^{rd}$, $6^{th}$, $7^{th}$ and $8^{th}$ column. Furthermore, as shown in the $1^{st}$, $2^{nd}$, $4^{th}$, $6^{th}$ and $8^{th}$ column, Fusion-S2iGan produces visually plausible birds with more vivid details than AttnGAN and S2IGAN.  

The qualitative results of DTGAN \cite{zhang2021dtgan}, S2IGAN \cite{wang2021generating} and Fusion-S2iGan on the Oxford-102 data set are illustrated in Fig. \ref{q1}(b), suggesting that Fusion-S2iGan is able to adopt a single generator/discriminator pair to produce high-resolution pictures that correspond well to the given speech-audio descriptions. For instance, Fusion-S2iGan yields perceptually realistic flowers with a more vivid shape than DTGAN and S2IGAN in the $1^{st}$, $2^{nd}$, $6^{th}$, $7^{th}$ and $8^{th}$ column. In addition, as shown in the $3^{th}$, $4^{th}$, $5^{th}$ and $6^{th}$ column, the flowers generated by Fusion-S2iGan have more clear details and richer color distributions than DTGAN and S2IGAN. 

The above experimental results indicate that Fusion-S2iGan equipped with the dual-residual speech-visual fusion blocks has the ability to effectively inject the speech information into the generator network and generate high-quality photographs.
\section{Results on the real speech datasets}
\label{s7}
In addition to the synthesized speech data sets, we also evaluate Fusion-S2iGan on the challenging real speech data sets, i.e., Flickr8k and Places-subset data sets, from both quantitative and qualitative perspectives. 
\subsection{Quantitative Results}
Table  \ref{t2} reports the IS, FID and R@50 of Fusion-S2iGan and other compared approaches on the Flickr8k and Places-subset data sets. It can be observed that Fusion-S2iGan performs better than S2IGAN by notably enhancing the IS from 8.72 to 11.70 on the Flickr8k data set and from 4.04 to 5.05 on the Places-subset data set, reducing the FID from 93.29 to 70.80 on the Flickr8k data set and from 42.09 to 25.68 on the Places-subset data set and improving the R@50 from 16.40 to 34.95 on the Flickr8k data set and from 12.95 to 28.37 on the Places-subset data set. Notably, Fusion-S2iGan obtains a remarkably lower FID in speech-to-image synthesis than AttnGAN doing text-to-image generation on both data sets, which demonstrates that the distribution of the samples synthesized by our model is closer to the real data distribution. Specifically, Fusion-S2iGan reduces the FID from 84.08 to 70.80 on the Flickr8k data set and from 35.59 to 25.68 on the Places-subset data set while increasing the IS from 4.59 to 5.05 on the Places-subset data set. We can also see that Fusion-S2iGan gets a lower R@50 score than AttnGAN on both data set. The reason for this may be that speech signals lack regular breaks between words, i.e., the spaces in the written-form text.
\begin{table}
\caption{The IS, FID and R@50 of AttnGAN \cite{xu2018attngan} based on the textual descriptions, Li \textit{et al.}, StackGAN++ \cite{zhang2018stackgan++}, S2IGAN \cite{wang2021generating} and Fusion-S2iGan conditioned on the spoken captions on the Flickr8k and Places-subset data sets. The best results of speech-to-image generation are in bold.}
\begin{center}
\scalebox{0.88}{
\begin{tabular}{c c c c c c}
\midrule
Datasets & Methods & Input & IS $\uparrow$ & FID $\downarrow$ & R@50 $\uparrow$\\
\midrule
\multirow{5}*{Flickr8k} 
& AttnGAN \cite{xu2018attngan} & text & 12.37$\pm$0.41 & 84.08 & 50.40\\
\cmidrule(r){2-6}
& StackGAN++ \cite{zhang2018stackgan++} & speech & 8.36 $\pm$0.39 & 101.74 & 16.40\\
& S2IGAN \cite{wang2021generating} & speech & 8.72$\pm$0.34 & 93.29 & 16.40\\
\cmidrule(r){2-6}
& \textbf{Fusion-S2iGan} & speech & \textbf{11.70$\pm$0.45} & \textbf{70.80} & \textbf{34.95}\\
\midrule
\multirow{6}*{Places-subset}
& AttnGAN \cite{xu2018attngan} & text & 4.59$\pm$0.51 & 35.59 & 33.85\\
\cmidrule(r){2-6}
& Li \textit{et al.} & speech & $-$ & 83.06 & $-$ \\
& StackGAN++ \cite{zhang2018stackgan++} & speech & 3.78$\pm$0.35 & 47.94 & 8.87 \\
& S2IGAN \cite{wang2021generating} & speech & 4.04$\pm$0.25 & 42.09 & 12.95\\
\cmidrule(r){2-6}
& \textbf{Fusion-S2iGan} & speech & \textbf{5.05$\pm$0.10} & \textbf{25.68} & \textbf{28.37}\\
\midrule
\end{tabular}}
\end{center}
\label{t2}
\vspace{-0.1in}
\end{table}
\begin{figure*}[t]
   \begin{minipage}[b]{1.0\linewidth}
   \centerline{\includegraphics[width=180mm]{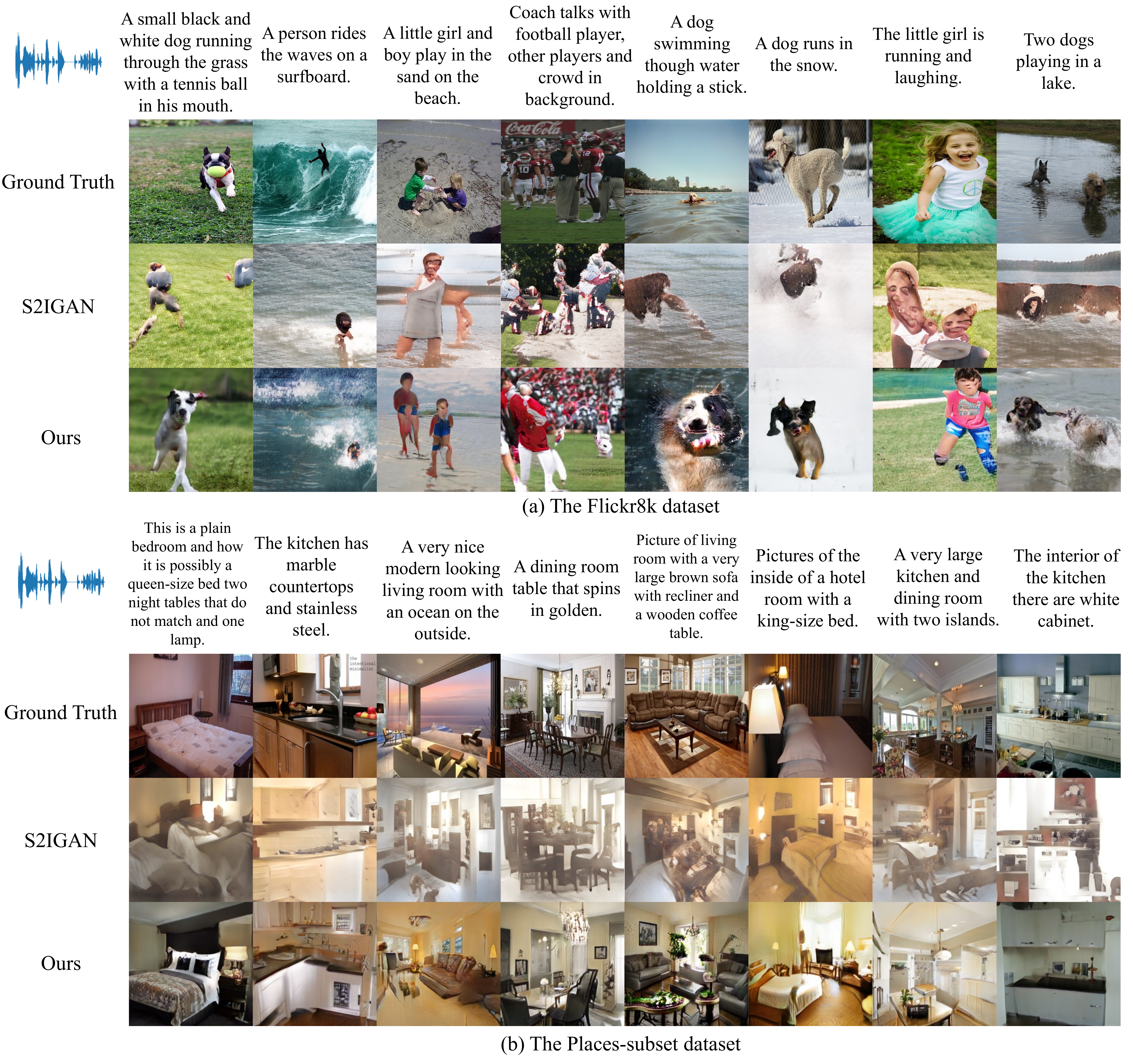}}
   \end{minipage}
   \caption{Qualitative comparison of S2IGAN \cite{wang2021generating} and Fusion-S2iGan conditioned on the speech signals on the Flickr8k and Places-subset data sets. Ground-truth images are shown above the samples produced by S2IGAN and the spoken descriptions are presented above the ground-truth images.}
   \vspace{-0.1in}
  \label{q2} 
\end{figure*}
\subsection{Qualitative Results}
For qualitative comparison, we visualize the pictures produced by Fusion-S2iGan and S2IGAN and corresponding ground-truth images on the Flickr8k and Places-subset data sets in Fig. \ref{q2}, which suggests that Fusion-S2iGan is capable of yielding photo-realistic and semantic-consistency photographs when given speech signals. For example, in terms of complex-scene generation, Fusion-S2iGan generates visually plausible dogs with more vivid details and more clear backgrounds than S2IGAN in the $1^{st}$, $5^{th}$, $6^{th}$ and $8^{th}$ column. We can also observe that Fusion-S2iGan synthesizes a man surfing on the realistic sea waves ($2^{nd}$ column), two clear persons standing on the beach ($3^{rd}$ column), a plausible football player in red and a reasonable background ($4^{th}$ column) and a running girl ($4^{th}$ column), whereas S2IGAN yields unclear objects ($1^{st}$, $2^{nd}$, $4^{th}$, $5^{th}$, $6^{th}$ and $8^{th}$ column) and blurry backgrounds ($4^{th}$ and $7^{th}$ column). Furthermore, it can be seen that the number of the objects produced by Fusion-S2iGan is correct in the $1^{st}$, $2^{nd}$, $3^{rd}$, $5^{th}$, $6^{th}$, $7^{th}$ and $8^{th}$ column. 

As can be seen in Fig. \ref{q2}(b), Fusion-S2iGan produces promising and high-quality complex-scene pictures, i.e., a realistic bedroom ($1^{st}$ column), plausible kitchens ($2^{nd}$ and $8^{th}$ column), high-quality living rooms ($3^{rd}$ and $5^{th}$ column), visually promising dining rooms ($4^{th}$ and $7^{th}$ column) and a photo-realistic hotel room ($6^{th}$ column), although the Places-subset data set is very challenging.  However, some examples generated by S2IGAN ($1^{st}$, $4^{th}$, $6^{th}$, $7^{th}$ and $8^{th}$ column) are not plausible. For instance, the shape of the beds is not clear ($1^{st}$ and $6^{th}$ column) and the color distribution is rough ($4^{th}$ and $8^{th}$ column). 

The above analysis indicates that Fusion-S2iGan has the capacity to achieve very good results on the real speech data sets due to the effective speech-visual fusion module and loss functions.  
\section{ablation/substitution tests}
\label{s8}
In order to evaluate the effectiveness of different components in Fusion-S2iGan, we perform a series of ablation/substitution tests on the CUB, Oxford-102 and Flickr8k and Places-subset data sets. Ablation usually refers to the removal of processing steps and evaluating the effects on performance. Here, however, the word `substitution' is more appropriate, referring to the in-place replacement of a network module by another variant. In the evaluation we used 30k sample images for the CUB data set and 10k samples for the Oxford-102 data set in order to obtain reliable performance estimates.

\subsection{Effectiveness of the Pixel-Attention Module (PAM)}
To validate the effectiveness of the introduced pixel-attention module (PAM), we investigate the performance of Fusion-S2iGan with other attention modules on the CUB and Oxford-102 data sets. Specifically, we replace the PAM in the visual+speech fusion module (VSFM) using the bottleneck attention module (BAM) \cite{park2018bam}, polarized self-attention (PolarizedAttn) \cite{liu2021polarized} and contextual transformer attention (CoTAttn) \cite{li2022contextual}, respectively. The comparison results are depicted in Table \ref{t3}. We can see that PAM outperforms BAM, PolarizedAttn and CoTAttn by increasing the IS by 0.12 on the CUB data set and 0.06 on the Oxford-102 data set, reducing the FID by 0.33 on the CUB data set and 1.16 on the Oxford-102 data set and enhancing the mAP by 0.75 on the CUB data set and 0.44 on the Oxford-102 data set. The results demonstrate that PAM can improve the quality and semantic relevancy of produced pictures, relative to the other methods. 
\begin{table}
\caption{Substitution study on pixel-attention modules. BAM, PolarizedAttn and CoTAttn refer to the bottleneck attention module \cite{park2018bam}, polarized self-attention \cite{liu2021polarized} and contextual transformer attention \cite{li2022contextual}, respectively. PAM indicates the introduced pixel-attention module. The best results are in bold.}
\begin{center}
\scalebox{0.85}{
\begin{tabular}{l c c c c c c}
\midrule
Datasets & \multicolumn{3}{c}{CUB} & \multicolumn{3}{c}{Oxford-102}\\
\cmidrule(r){2-4}  \cmidrule(r){5-7} 
Evaluation Metrics & IS $\uparrow$ & FID $\downarrow$ & mAP $\uparrow$ & IS $\uparrow$ & FID $\downarrow$ & mAP $\uparrow$\\  
\midrule
BAM \cite{park2018bam} &  4.35$\pm$0.04 & 16.23 & 10.32  &  3.62$\pm$0.09 & 42.72 & 16.35 \\
PolarizedAttn \cite{liu2021polarized} &  4.68$\pm$0.04 & 14.33 & 10.74  & 3.75$\pm$0.07  & 44.20 & 16.42 \\
CoTAttn \cite{li2022contextual} &  4.70$\pm$0.05 &  14.07 & 10.53  & 3.75$\pm$0.06 & 41.24 & 17.28 \\
\midrule
\textbf{PAM} & \textbf{5.06$\pm$0.09} & \textbf{13.09} & \textbf{12.12} & \textbf{3.81$\pm$0.08} & \textbf{40.08} & \textbf{17.72}\\
\midrule
\end{tabular}}
\end{center}
\label{t3}
\vspace{-0.1in}
\end{table}
\subsection{Effectiveness of the Weighted-Fusion Module (WFM)}
To further prove the benefits of the proposed weighted-fusion module (WFM), we conduct a substitution test on fusion methods. We replace WFM with the element-wise addition and multiplication operations, respectively. Table \ref{t4} reports the quantitative results on the CUB and Flickr8k data sets. It can be observed that WFM performs better than both addition and product operations, significantly improving the IS from 4.68 to 4.82 on the CUB data set and from 10.23 to 11.70 on the Flickr8k data set, reducing the FID from 15.25 to 13.74 on the CUB data set and from 78.03 to 70.80 on the Flickr8k data set and increasing the mAP from 11.11 to 11.49 on the CUB data set and the R@50 from 30.82 to 30.95 on the Flickr8k data set. The above analysis suggests the effectiveness of the presented WFM in comparison to plain additive or product-based weighing.
\begin{table}
\caption{Substitution test on fusion methods. Addition and Product represent the element-wise addition and product operation, respectively. WFM denotes the proposed weighted-fusion module. The best results are in bold.}
\begin{center}
\scalebox{0.8}{
\begin{tabular}{l c c c c c c}
\midrule
Datasets & \multicolumn{3}{c}{CUB} & \multicolumn{3}{c}{Flickr8k}\\
\cmidrule(r){2-4}  \cmidrule(r){5-7} 
Evaluation Metrics & IS $\uparrow$ & FID $\downarrow$ & mAP $\uparrow$ & IS $\uparrow$ & FID $\downarrow$ & R@50 $\uparrow$\\  
\midrule
Addition & 3.70$\pm$0.04 & 29.11 & 2.7 & 9.53$\pm$0.37 & 100.83 & 5.98  \\
Product & 4.68$\pm$0.04 & 15.25 & 11.11  & 10.23$\pm$0.36 & 78.03 & 30.82 \\
\midrule
\textbf{WFM} & \textbf{5.06$\pm$0.09} & \textbf{13.09} & \textbf{12.12} & \textbf{11.70$\pm$0.45} & \textbf{70.80} & \textbf{34.95}\\
\midrule
\end{tabular}}
\end{center}
\label{t4}
\vspace{-0.1in}
\end{table}
\subsection{Effectiveness of the Visual+Speech Fusion Module (VSFM)}
\label{e_vsfm}
To verify the effectiveness of the developed visual+speech fusion module (VSFM) and investigate how far can current single-stage text-to-image algorithms be used for speech-to-image generation, we explore the results of Fusion-S2iGan with the sentence-visual fusion modules in DTGAN \cite{zhang2021dtgan} and DFGAN \cite{tao2020df} on the CUB and Oxford-102 data sets. The results of the substitution study are displayed in Table \ref{t5}. We can see that the VSFM achieves the best score, notably increasing the IS by 0.42 on the CUB data set and 0.23 on the Oxford-102 data set, reducing the FID by 3.00 on the CUB data set and 8.29 on the Oxford-102 data set and increasing the mAP by 3.29 on the CUB data set and 1.40 on the Oxford-102 data set. 

It can also be observed that DTGAN does not obtain promising performance on the speech data sets. The reason behind this result may be that the dual-attention module in DTGAN is not suitable for processing the continuous speech signals and modeling the semantic relationships between vision and speech. The above experimental results indicate that with the VSFM, Fusion-S2iGan is able to yield high-resolution pictures semantically aligning with the input speech-audio descriptions. 

To evaluate the impact of different number of the fusion modules in the dual-residual speech-visual fusion block, we compare the results of a fusion module (single-module) and two fusion modules (dual-module) on the Flickr8k and Places-subset data sets, as shown in Table \ref{t6}. We can observe that the proposed dual-residual speech-visual fusion block outperforms the block with a single module on both data set, which validates the effectiveness of the dual-residual structure. 
\begin{table}
\caption{Substitution study on the visual+speech fusion module (VSFM). DTGAN and DF-GAN refer to the sentence-visual fusion module in DTGAN \cite{zhang2021dtgan} and DF-GAN \cite{tao2020df}, respectively. VSFM denotes the presented visual+speech fusion module. The best results are in bold.}
\begin{center}
\scalebox{0.8}{
\begin{tabular}{l c c c c c c}
\midrule
Datasets & \multicolumn{3}{c}{CUB} & \multicolumn{3}{c}{Oxford-102}\\
\cmidrule(r){2-4}  \cmidrule(r){5-7} 
Evaluation Metrics & IS $\uparrow$ & FID $\downarrow$ & mAP $\uparrow$ & IS $\uparrow$ & FID $\downarrow$ & mAP $\uparrow$\\
\midrule
DTGAN \cite{zhang2020dtgan} & 4.18 $\pm$0.04 &  24.70 &  4.92 & 3.15$\pm$0.04 &  52.67 &  13.87  \\
DF-GAN \cite{tao2020df} & 4.40 $\pm$0.03 &  16.74 &  8.20  & 3.58$\pm$0.07 &  48.37 &  16.32 \\
\midrule
\textbf{VSFM} & \textbf{5.06$\pm$0.09} & \textbf{13.09} & \textbf{12.12} & \textbf{3.81$\pm$0.08} & \textbf{40.08} & \textbf{17.72}\\
\midrule
\end{tabular}}
\end{center}
\label{t5}
\vspace{-0.1in}
\end{table}
\begin{table}[t]
\caption{Effect of the number of fusion modules in the dual-residual speech-visual fusion block. Single-module and dual-module indicate the residual structure with a single fusion module and two fusion modules, respectively. The best scores are in bold.}
\begin{center}
\scalebox{0.8}{
\begin{tabular}{l c c c c c c}
\midrule
Datasets & \multicolumn{3}{c}{Flickr8k} & \multicolumn{3}{c}{Places-subset}\\
\cmidrule(r){2-4}  \cmidrule(r){5-7} 
Evaluation Metrics & IS $\uparrow$ & FID $\downarrow$ & mAP $\uparrow$ & IS $\uparrow$ & FID $\downarrow$ & mAP $\uparrow$\\  
\midrule
single-module &  10.39$\pm$0.34 & 81.62 & 32.37 & 5.02$\pm$0.06 & 41.67 & 24.00 \\
\textbf{dual-module} & \textbf{11.70$\pm$0.45} & \textbf{70.80} & \textbf{34.95} & \textbf{5.05$\pm$0.10} & \textbf{25.68} & \textbf{28.37}\\
\midrule
\end{tabular}}
\end{center}
\label{t6}
\vspace{-0.1in}
\end{table}
\section{Conclusion}
\label{s9}
In this paper, we propose a unified, novel end-to-end speech-to-image transform architecture named Fusion-S2iGan. Fusion-S2iGan is able to only use a generator/discriminator pair to project a speech signal into a high-quality picture directly. Fusion-S2iGan adopts a new and effective visual+speech fusion module (VSFM) to modulate the visual feature maps with the speech information and boost the resolution of produced samples. More significantly, Fusion-S2iGan spreads the bimodal information over all layers of the generative model. This allows for an influence of the speech over features at various hierarchical levels in the architecture, from crude early features to abstract late features. The hinge objective, deep attentional multimodal similarity model (DAMSM) loss and matching-aware zero-centered gradient penalty (MA-GP) loss are introduced to stabilize the training of the conditional generative adversarial network (cGAN). Fusion-S2iGan is evaluated on four public data sets, i.e., CUB bird, Oxford-102, Flickr8k and Places-subset, outperforming current speech-to-image methods. Moreover, we explore whether current text-to-image approaches are effective for a speech-to-image transform. Furthermore, our presented VSFM is a general fusion module, and can be easily integrated into current speech-to-image frameworks to improve image quality and semantic consistency. More significantly, our developed architecture overcomes the issues presenting in the existing multi-stage speech-to-image algorithms, and can serve as a strong basis for developing better speech-to-image generation models. In the future, we will investigate how to inject the emotional features derived from the speech signal into the speech-to-image synthesis pipeline and how to combine speech signals with other conditional contexts, e.g., sketches, to control the produced pictures.


{\small
\bibliographystyle{IEEEtran}
\bibliography{egbib}
}


\vfill

\end{document}